\documentclass[lettersize,journal]{IEEEtran}
\usepackage{amsmath,amsfonts}
\usepackage{algorithmic}
\usepackage{algorithm}
\usepackage{array}
\usepackage[caption=false,font=normalsize,labelfont=sf,textfont=sf]{subfig}
\usepackage{textcomp}
\usepackage{stfloats}
\usepackage{url}
\usepackage{verbatim}
\usepackage{graphicx}
\usepackage{cite}
\usepackage{tikz}
\usepackage{comment}
\usepackage{amsmath,amssymb} 
\usepackage{color}
\usepackage{orcidlink}
\usepackage{booktabs}
\usepackage[accsupp]{axessibility}  
\usepackage{threeparttable}
\usepackage{bbding}
\usepackage{multirow}
\usepackage{wrapfig}
\definecolor{lemon}{RGB}{240,220,40}

\begin{document}
\title{ConSept: Continual Semantic Segmentation via Adapter-based Vision Transformer}

\author{Bowen Dong$^{1,2}$, Guanglei Yang$^{1}$, Wangmeng Zuo$^{1}$\textsuperscript{\Envelope}, Lei Zhang$^{2}$
\thanks{1. School of Computer Science and Technology, Harbin
Institute of Technology, Harbin 150001, China (e-mail: cswmzuo@gmail.com).
2. Department of Computing, the Hong Kong Polytechnic University: (e-mail: cslzhang@comp.polyu.edu.hk). \Envelope \quad denotes corresponding author.}}

\markboth{Preprint. Work in progress.}
{Shell \MakeLowercase{\textit{et al.}}: A Sample Article Using IEEEtran.cls for IEEE Journals}


\maketitle

\begin{abstract}
%
In this paper, we delve into the realm of vision transformers for continual semantic segmentation, a problem that has not been sufficiently explored in previous literature. 
%
%
Empirical investigations on the adaptation of existing frameworks to vanilla ViT reveal that 
incorporating visual adapters into ViTs or fine-tuning ViTs with distillation terms is advantageous for enhancing the segmentation capability of novel classes.
%
These findings motivate us to propose \emph{Con}tinual semantic \emph{Se}gmentation via Ada\emph{pt}er-based ViT, namely \emph{ConSept}.
Within the simplified architecture of \emph{ViT with linear segmentation head}, ConSept integrates lightweight attention-based adapters into vanilla ViTs.
%
Capitalizing on the feature adaptation abilities of these adapters, ConSept not only retains superior segmentation ability for old classes, but also attains promising segmentation quality for novel classes. 
To further harness the intrinsic anti-catastrophic forgetting ability of ConSept and concurrently enhance the segmentation capabilities for both old and new classes, we propose two key strategies: 
distillation with a deterministic old-classes boundary for improved anti-catastrophic forgetting, 
and dual dice losses to regularize segmentation maps, thereby improving overall segmentation performance.
%
%
Extensive experiments show the effectiveness of ConSept on multiple continual semantic segmentation benchmarks under \emph{overlapped} or \emph{disjoint} settings. 
Code will be publicly available at \url{https://github.com/DongSky/ConSept}.
\end{abstract}

\section{Introduction}\label{sec:introduction}
Semantic segmentation~\cite{everingham2010pascal,lin2014microsoft,chen2017deeplab} aims to recognize and segment region masks for the given categories. 
The evolution of semantic segmentation algorithms~\cite{chen2017deeplab,cheng2021mask2former,strudel2021segmenter,xie2021segformer,xiao2018unified} has allowed the segmentation models to produce precise masks for individual images. However, in practical scenarios, these aforementioned algorithms are expected to possess the capability to assimilate novel concepts continually through specific learning paradigms, \emph{i.e.}, continual semantic segmentation, which ensures that the acquired algorithms exhibit proficient performance across both old and newly added categories and mitigate the phenomenon of catastrophic forgetting~\cite{french1999catastrophic,li2016learning}. In this paper, we focus on the class-incremental setting~\cite{cermelli2020modeling,wang2022learning,wang2022dualprompt} and delve into discussions regarding methods for continual semantic segmentation.

Generally speaking, prior networks dedicated to continual semantic segmentation~\cite{cha2021ssul,zhang2022microseg,cermelli2020modeling,douillard2021plop} commonly employ a foundational learner constructed through the utilization of  a CNN-based feature extractor~\cite{He2016DeepRL} and a DeepLab segmentation head~\cite{chen2017deeplab}. Then these learners are accompanied through strategies such as a) knowledge distillation methods~\cite{cermelli2020modeling,douillard2021plop}, b) regularization-based methods~\cite{michieli2021continual,lin2022continual}, and c) exemplar-replay methods~\cite{zhu2023continual,yan2021framework,maracani2021recall}. These frameworks notably enhance the learner's resilience against catastrophic forgetting. However, existing CNN-based methods encounter two fundamental challenges. Firstly, the fixed-size convolution kernels impede the long-range interaction capacity of feature extractors, thereby constraining the overall segmentation performance. Secondly, the anti-catastrophic forgetting ability to old classes remains constrained, thereby limiting the applicability of continual semantic segmentation methods.

\begin{figure}[t]
\begin{center}
\includegraphics[width=0.48\textwidth]{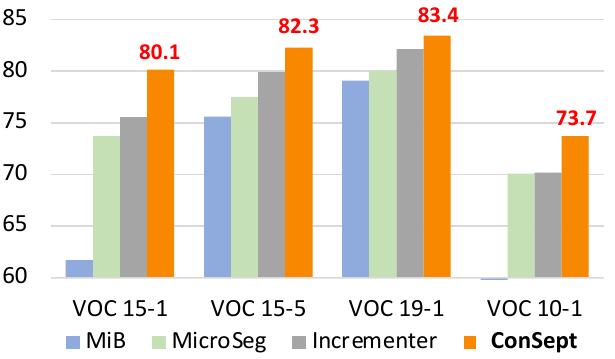}
\end{center}
\vspace{-1em}
\caption{
    Performance comparison between ConSept and state-of-the-art methods~\cite{cermelli2020modeling,zhang2022microseg,shang2023incrementer}. ConSept obtains the best performance on all PASCAL VOC benchmarks with \emph{overlapped} setting. Best viewed in color.
}
\vspace{-1em}
\label{fig:overview_voc}
\end{figure}

Inspired by state-of-the-art Vision Transformer (ViT)-based segmentation methods~\cite{SETR,strudel2021segmenter,xie2021segformer}, one can utilize the ViT feature extractor to construct ViT-based continual semantic segmentation frameworks~\cite{shang2023incrementer,zhang2023coinseg,cermelli2023comformer} for better performance. However, existing state-of-the-art ViT-based continual segmenters  face several critical issues, including the reliance on heavy segmentation decoders~\cite{shang2023incrementer,cermelli2023comformer,zhang2023coinseg}, and the need of extra region proposals~\cite{zhang2023coinseg} to maintain optimal performance. Such drawbacks limit the practical applications of such methods. Consequently, a question arises: \emph{could we formulate an effective continual segmentation framework using vision transformers with straightforward linear segmentation heads}?

To answer this question, we start from a pretrained ViT~\cite{dosovitskiy2020image} with linear segmentation head to conduct preliminary study. 
Upon integrating such a model into a classical CNN-based continual semantic segmentation method (\emph{e.g.}, SSUL~\cite{cha2021ssul}), 
the ViT-based continual segmenter exhibits comparable performance on base classes but experiences a notable decline in mIoU on novel classes, consequently resulting in an overall performance degeneration. 
Through empirical investigation into pretrained vision transformers, it is discerned that a frozen pretrained vision transformer tends to overfit on base classes, 
thereby preserving feasible anti-catastrophic forgetting capability on these classes while concurrently constraining its generalization capacity to novel categories. 
Furthermore, a comprehensive empirical analysis in Sec.~\ref{sec:pre_analysis} reveals that incorporating adapters into the ViT feature extractor and introducing additional distillation terms are advantageous for both old and novel classes in the context of continual semantic segmentation.
These findings naturally motivate us to design an adapter-based vision transformer and integrate it with specifically designed continual learning frameworks. 
Such an architecture, even in the absence of complex segmentation decoders, could accomplish the dual objectives of: 
\textbf{1)} enhancing anti-catastrophic forgetting ability for base classes, 
and \textbf{2)} attaining robust segmentation capability for novel classes. 
With the above motivations, we present ConSept, a straightforward yet efficacious continual semantic segmentation method. 

Maintaining a simplistic macro architecture comprising solely ViT~\cite{dosovitskiy2020image} and a linear segmentation head, 
ConSept sheds light on ViT-based continual semantic segmentation.
A primary concern on freezing ViT lies in its sub-optimal segmentation performance on novel categories. 
In response to this challenge, we propose a pivotal strategy, namely \textbf{fine-tuning with adapters for better generalization}, which is the most significant component of ConSept. 
Inspired by prior works on visual adapters~\cite{nie2022protuning,chen2022convadapter,chen2023vision} for transfer learning, we integrate a shallow convolution-based stem block and lightweight attention-based adapters into the pretrained ViT~\cite{dosovitskiy2020image}, resulting in a dual-path feature extractor. 
The image features refined by these adapters are subsequently input into a simple linear segmentation head for prediction. 
With less than $10\%$ additional parameters, ConSept notably enhances the generalization ability for novel classes, preserving the anti-catastrophic forgetting capability for base classes effectively and efficiently.
Note that conventional approaches~\cite{cha2021ssul,zhang2022microseg} often adopt the strategy of freezing feature extractors during training to mitigate the risk of catastrophic forgetting. 
However, in the context of ViT-based scenarios shown in Table~\ref{table:pre_vanilla_vit}, this strategy imposes substantial limitations on the segmentation ability of novel classes. 
To mitigate the adverse effects associated with frozen ViT and leverage the benefits of adapters within ConSept, we advocate fine-tuning the entire feature extractor. 
This strategic adjustment ensures the network's capacity to learn representative features for new concepts,  reinforcing its generalization ability on novel classes.

In conjunction with the adapter-based macro design in ConSept, we introduce two additional strategies to further harness the efficacy of adapter-based ConSept for continual segmentation. 
The first strategy involves \textbf{distillation with deterministic old-classes boundary for better anti-catastrophic forgetting}. 
Since directly fine-tuning ViT feature extractor without any constraints could still lead to catastrophic forgetting for base classes~\cite{shang2023incrementer,cermelli2020modeling}, to further enhance the anti-forgetting capability, we propose to employ dense mean-square loss and dense contrastive loss to distill image features between old and new models, and preserve the frozen state of the linear head for old classes to uphold a well-defined decision boundary for segmentation.
The second strategy revolves the \textbf{dual dice losses for better regularization}. 
Specifically, we incorporate the two dice losses to assess errors in segmentation predictions. 
For the class-specific dice loss term, we adhere to the classical dice loss~\cite{sudre2017generalised} to enhance overall segmentation proficiency. 
Meanwhile, for the old-new dice loss term, we binarize the pseudo ground-truth maps into regions corresponding to old classes and new classes, and apply dice loss constraint to improve the segmentation quality for novel classes.

Experiments are conducted on two challenging continual semantic segmentation benchmarks, \emph{i.e.}, PASCAL VOC~\cite{everingham2010pascal} and ADE20K~\cite{zhou2017scene}. 
As shown in Fig.~\ref{fig:overview_voc}, 
ConSept achieves leading performance on PASCAL VOC benchmarks under various class-incremental learning steps. 
Even in the challenging scenarios with more classes, 
ConSept can still obtain remarkable segmentation performance for both base and novel classes. 
The promising results demonstrate that, 
\emph{albeit without relying on extra object proposals from external models~\cite{zhang2022microseg,zhang2023coinseg} or heavy segmentation decoder~\cite{shang2023incrementer}}, 
ConSept can still enhance the anti-catastrophic forgetting capability of base classes, 
meanwhile effectively improve the segmentation ability for novel classes. 
Our method provides a strong and stable baseline for ViT-based continual semantic segmentation tasks.
\section{Related Work}\label{sec:related_work}
\subsection{Continual Learning}\label{sec:rel_continual}
Deep neural networks with the vanilla \emph{``pre-training then finetuning'' paradigm} often face the problem of severe catastrophic forgetting~\cite{french1999catastrophic} in various tasks (\emph{e.g.}, image classification), resulting in unsatisfying performance on old categories. 
To tackle this issue, continual learning approaches~\cite{li2016learning} have been proposed to maintain the recognition ability on old classes, meanwhile obtain promising recognition accuracy on novel categories. 
Specifically, Li \emph{et al.} proposed LwF~\cite{li2016learning}, which builds the consistency between old models and new models on confidence scores of seen categories, thus reducing the effect of forgetting during novel classes training. 
Further works follow the fundamental design of LwF, and manage to solve catastrophic forgetting in four classes of methods. 
The first class is distillation-based methods, which conduct knowledge distillation on feature representation~\cite{douillard2020podnet,kang2022class} or output confidence scores~\cite{li2016learning} between old and new models. 
The second class is replay-based methods~\cite{rebuffi2017icarl,kang2022class}, which store a small number of exemplars from seen categories, and utilize them during the training of newly-added tasks to avoid catastrophic forgetting explicitly.
The third class is regularization-based methods~\cite{joseph2022energy}, which introduce explicit constraints on model parameters between two tasks or design implicit regularization on feature representations to avoid forgetting.
The last class is architecture-based methods~\cite{li2019learn}, which mainly focus on fine-tuning specially designed new layers or adapters to obtain promising performance on new tasks and maintain accuracy on old tasks.
Our work can be  categorized into distillation-based method. Different from the above works, we focus on the more challenging continual semantic segmentation task.

\subsection{Continual Semantic Segmentation} \label{sec:rel_conseg}
Semantic segmentation~\cite{long2015fully,xiao2018unified,strudel2021segmenter,xie2021segformer} aims to identify the category for each pixel in an image. To meet the need of continual learning, Michieli \emph{et al.}~\cite{michieli2019incremental} started from DeepLab~\cite{chen2017deeplab} with CNN-based backbone~\cite{He2016DeepRL} and proposed the basic structure of continual semantic segmentation framework via knowledge distillation. Methods of continual semantic segmentation can be also divided into three categories. 
The first category is distillation-based methods~\cite{cermelli2020modeling,douillard2021plop}.  Cermelli \emph{et al.}~\cite{cermelli2020modeling} proposed the continual semantic segmentation task as well as the baseline method namely MiB. During training of new tasks, MiB reconstructs the confidence score of background class from newly-added classes to maintain the consistency between old and new models to avoid catastrophic forgetting. Douillard \emph{et al.} proposed PLOP~\cite{douillard2021plop}, which extends POD feature~\cite{douillard2020podnet} by using the spatial-pyramid scheme, obtaining better performance on base classes. 
The second category is regularization-based methods~\cite{michieli2021continual,lin2022continual}. For example, Michieli \emph{et al.} proposed SDR~\cite{michieli2021continual}, which leverages prototype matching to enforce latent space consistency on old classes, thus avoiding catastrophic forgetting. Lin \emph{et al.} proposed the structure preserving loss to maintain inter-class structure and intra-class structure, respectively.
The third category is replay-based methods~\cite{zhu2023continual,yan2021framework,maracani2021recall}. Similar to replay-based continual image classification, these methods store a small number of representative samples with old classes annotation, and replay these samples during new tasks training. 

Nevertheless, though vision transformers have shown remarkable performance on dense prediction tasks, only a few works~\cite{cermelli2023comformer,shang2023incrementer} have managed to investigate transformer-based continual segmentation. All these methods largely leverage specially-designed transformer-based decoders~\cite{strudel2021segmenter,cheng2021mask2former} to obtain promising performance. 
Our work also focuses on continual semantic segmentation with ViTs. Different from previous works, our work illustrates that, even without heavy decoder for accurate segmentation, we can still ensure high prediction quality for both old and new categories.

\subsection{Vision Transformers} \label{sec:rel_vit}
Benefiting from stacked self-attention mechanisms~\cite{vaswani2017attention} and feed-forward networks, vision transformers (ViTs)~\cite{dosovitskiy2020image,touvron2021training} have demonstrated remarkable performance on various vision tasks~\cite{detr20,YOLOS,strudel2021segmenter,xie2021segformer,dong2022self,dong2023lpt,radford2021learning}. 
Conventional exploration of vision transformers focus on two paradigms. 
The first is the classical ``large-scale pre-training then fine-tuning'' paradigm~\cite{detr20,YOLOS,xie2021segformer,strudel2021segmenter,Zheng_2021_CVPR,dong2023lpt}. 
This paradigm starts from ImageNet-pretrained~\cite{deng2009imagenet,caron2021emerging,he2021masked} vision transformers and then conducts full fine-tuning~\cite{strudel2021segmenter,xie2021segformer,detr20,touvron2021training} or parameter-efficient tuning~\cite{jia2022vpt,chen2022convadapter,dong2023lpt} on standard downstream tasks (\emph{e.g.}, detection~\cite{lin2014microsoft} or segmentation~\cite{zhou2017scene}), thus obtaining desired performance on corresponding tasks. 
The second paradigm is transformer-based few-shot learning~\cite{he2022attribute,dong2022self,chen2023semantic}, which focuses on learning robust and generalized image feature representations, such that the learned ViTs are able to recognize novel categories without adaptation or with only fast low-shot adaptation. 
The above works mainly concern the performance of newly-added tasks while ignoring how to evaluate the capability against catastrophic forgetting. 
Recent works have explored continual learning with vision transformers in conventional continual image classification settings~\cite{yu2021improving} or learning with pretrained models settings~\cite{wang2022learning,wang2022dualprompt,smith2023coda,jeeveswaran2023birt,Mohamed_2023_CVPR}.
%
%
Different from previous works, we aim to explore the continual learning capability of ViTs on the more challenging downstream segmentation tasks~\cite{cermelli2023comformer,shang2023incrementer}.
Our experimental results illustrate that, even without complicated segmentation heads~\cite{strudel2021segmenter,chen2017rethinking,xiao2018unified}, we can still obtain promising results on multiple continual semantic segmentation benchmarks.
\section{Preliminary Study}\label{sec:preliminary}
\subsection{Problem Definition}\label{sec:pre_definition}
Building upon prior studies~\cite{michieli2021continual,cermelli2020modeling,shang2023incrementer,cha2021ssul,zhang2022microseg}, 
we focus on continual semantic segmentation tasks within class-incremental setting spanning $T$ steps. 
The definition of this task is provided below. 
Specifically, within the context of a semantic segmentation dataset $\mathcal{D}$ and its corresponding class set $\mathcal{C}$, 
we partition $\mathcal{C}$ into $T$ non-overlapping subsets of classes, denoted by $\{\mathcal{C}_{1}, ..., \mathcal{C}_{T}\}$.
Accordingly, the dataset $\mathcal{D}$ is partitioned into $T$ subsets $\{\mathcal{D}_{1}, ..., \mathcal{D}_{T}\}$, representing distinct tasks. 
For each task $t$ within $\mathcal{D}_{t}$, where $1\leq t \leq T$, the ground-truth segmentation masks corresponding to category $\mathcal{C}_{t}$ are accessible. 
It is conventionally assumed that $|\mathcal{C}_{2}| = \dots = |\mathcal{C}_{T}|$, yielding $X$-$Y$ continual semantic segmentation benchmarks, 
where $X = |\mathcal{C}_{1}|$ denotes the number of base classes and $Y = |\mathcal{C}_{2}|$ signifies the number of novel classes introduced in each new task.
The purpose of continual semantic segmentation is to learn a feature extractor $f$ as well as a segmentation decoder $g$ on each $\mathcal{D}_{t}$ sequentially (\emph{i.e.}, from $\mathcal{D}_{1}$ to $\mathcal{D}_{T}$) under the constraint that training annotations from $\mathcal{D}_{1:t-1}$ are not accessible. 
At each step $t>1$, the optimized $f$ with $g$ should obtain promising segmentation capability for both old categories $\mathcal{C}_{1:t-1}$ and newly learned categories $\mathcal{C}_{t}$. 

Distinguished from other tasks in visual recognition, the continual semantic segmentation task confronts two distinctive challenges. 
First, the complexity of accurately predicting pixel-level classification results exceeds that of obtaining image-level counterparts, 
and second, owing to the scarcity of abundant segmentation masks for $\mathcal{C}_{1:t-1}$ during training at step $t$, 
the segmentation network often faces pronounced catastrophic forgetting concerning $\mathcal{C}_{1:t-1}$. 
Consequently, the formulation of a straightforward yet efficient framework for continual semantic segmentation becomes highly demanding. 
Based on the remarkable progress in ViT-based dense prediction methods~\cite{xie2021segformer,strudel2021segmenter,SETR}, in this work we investigate in-depth how a vanilla ViT works on continual semantic segmentation. 

\begin{table*}[t]
    \centering
        \caption{Performance investigation of CNN-based and ViT-based continual semantic segmentation methods on the PASCAL VOC 15-1 task under \emph{overlapped} setting. 
        \textbf{(Upper)} Replacing the CNN backbone with vanilla ViT in the SSUL framework results in performance deterioration. 
        \textbf{(Lower)} Various ViT-based variants in continual segmentation. 
        Our findings indicate: \textbf{1)} an extremely lightweight linear decoder reduces forgetting on base classes; 
        \textbf{2)} fixing the backbone limits ViT's generalization for continual segmentation; 
        and \textbf{3)} introducing an adapter enhances anti-catastrophic forgetting and improves segmentation performance.
        }
        \label{table:pre_vanilla_vit}
        \vspace{-1em}
        \small
        \setlength{\tabcolsep}{3.5pt} 
       \renewcommand{\arraystretch}{4.0}
     { \fontsize{8.3}{3}\selectfont{
  
        \begin{tabular}{lccccc|ccc}
        \bottomrule
        \multirow{2}{*}{\textbf{Base Framework}}&\multirow{2}{*}{\textbf{Backbone}}&\multirow{2}{*}{\textbf{Decoder}}&\multirow{2}{*}{\textbf{Freeze}}&\multirow{2}{*}{\textbf{Distill}}&\multirow{2}{*}{\textbf{Adapter}}&\multicolumn{3}{c}{\textbf{VOC 15-1 (6 steps)}}
        \\ \cline{7-9}
        &&&&&&{{0-15}}&{{16-20}}&{all}
        \\ \hline
        SSUL~\cite{cha2021ssul} (Joint) & {ResNet-101}~\cite{He2016DeepRL} & DeepLab V3~\cite{chen2017deeplab} & \checkmark & - & -  &82.70 & 75.00 & 80.90 \\
        \hline
        SSUL~\cite{cha2021ssul} & {ResNet-101}~\cite{He2016DeepRL} & DeepLab V3~\cite{chen2017deeplab} & \checkmark & - & - &78.40 (-4.30\%) &49.00 (-26.00\%) &71.40 (-9.50\%)  \\
        SSUL~\cite{cha2021ssul} & ViT-B~\cite{dosovitskiy2020image}& DeepLab V3~\cite{chen2017deeplab} & \checkmark & - & -  &76.68 (-5.02\%)&43.78 (-31.22\%) &68.85 (-12.05\%) \\
        \bottomrule
        SSUL~\cite{cha2021ssul} & ViT-B~\cite{dosovitskiy2020image}& Linear & \checkmark & - & -  &80.81 (-1.89\%)& 26.87 (-48.13\%) & 67.97 (-12.93\%) \\
        SSUL~\cite{cha2021ssul} & ViT-B~\cite{dosovitskiy2020image}& Linear & - & - & -  &39.06 (-43.66\%)& 42.25 (-32.75\%) & 39.82 (-41.08\%) \\
        SSUL~\cite{cha2021ssul} & ViT-B~\cite{dosovitskiy2020image}& Linear & - & \checkmark & -  &80.43 (-2.27\%)& 62.67 (-12.33\%) & 76.20 (-14.70\%) \\
        SSUL~\cite{cha2021ssul} & ViT-B~\cite{dosovitskiy2020image}& Linear & \checkmark & - & \checkmark & {81.40 (-1.30\%)} & {53.40 (-21.60\%)} & {74.80 (-6.10\%)} \\

        \toprule
        \end{tabular}
        }}
        \vspace{-1em}
    \end{table*}
\subsection{Vanilla ViT on Continual Semantic Segmentation}\label{sec:investigation}
We are motivated by the following two factors to investigate vanilla ViT for continual semantic segmentation: 
\textbf{1)} the impressive performance demonstrated by pretrained vision transformers on segmentation tasks~\cite{strudel2021segmenter,xie2021segformer,SETR}; 
and \textbf{2)} the limited exploration of ViTs' anti-catastrophic forgetting capability in dense prediction scenarios. 
Given these considerations, ViTs are expected to outperform CNN-based approaches for continual semantic segmentation. 
We adopt the classical continual semantic segmentation framework SSUL~\cite{cha2021ssul} as the baseline method, 
which utilizes a frozen CNN feature extractor and the exemplar replay technique for optimal performance.
Specifically, we replace the original ResNet-101 feature extractor in SSUL with the ImageNet-pretrained ViT-B feature extractor~\cite{dosovitskiy2020image} and maintain other components unchanged. 
The baseline SSUL and the ViT-B variant are evaluated on the PASCAL VOC 15-1 benchmark under an \emph{overlapped} setting~\cite{everingham2010pascal,cermelli2020modeling}. 
The results are presented in the upper part of Table~\ref{table:pre_vanilla_vit}. 
Though introducing larger-range interactions and additional training parameters, the ViT-B variant does not surpass the baseline, achieving only 76.68\% mIoU on base classes, 1.72\% lower than its CNN counterpart. 
Compared to the CNN-based SSUL baseline, the ViT-B variant exhibits a significant performance decline by 5.22\% in mIoU on all novel classes, 
indicating challenges in learning new concepts in continual segmentation tasks. 
These findings prompt an investigation into the limitations and potential solutions for ViT-based continual semantic segmentation.

\subsection{Analysis}\label{sec:pre_analysis}
Based on prior ViT works~\cite{liu2021efficient,dong2022self}, if we perform  fine-tuning on $\mathcal{D}_{1}$, the segmentation network $f$ with decoder $g$ may overfit to base classes $\mathcal{C}_{1}$, limiting its generalization performance to novel categories $\mathcal{C}_{2:T}$ and resulting in lower overall mIoU. 
To test this hypothesis, we replace the DeepLab V3 segmentation head with a simple linear decoder, 
directly correlating predictions to ViT output features. 
In the lower part of Table~\ref{table:pre_vanilla_vit}, we see that introducing a linear head for the ViT-B variant yields a 4.13\% mIoU improvement on base classes, surpassing the CNN-based SSUL baseline by 2.4\%. 
This suggests that pretrained ViT possesses sufficient anti-catastrophic forgetting capability on base classes.
By removing the complex decoder structure to accommodate features for novel classes, the mIoU of novel classes notably drops to 26.87\%, 
underscoring the limited generalization performance of pretrained ViT to novel classes. This observation substantiates our hypothesis. 
Consequently, an intuitive question arises: \emph{can we employ the pretrained vanilla ViT with a linear decoder to formulate an effective continual segmentation network, 
achieving sustained strong segmentation for base classes and robust segmentation capability for novel classes?}

The answer to the above question unfolds in two aspects. 
\textbf{Firstly}, drawing inspiration from successful endeavors in parameter-efficient fine-tuning~\cite{hu2022lora,he2022towards,chen2022convadapter,nie2022protuning,chen2023vision}, 
we posit that \emph{incorporating lightweight modules, such as adapters~\cite{chen2022convadapter,chen2023vision,pmlr-v97-houlsby19a}, into the ViT feature extractor can enhance segmentation quality for novel classes}. 
To validate this hypothesis, we employ the ViT-B feature extractor with a linear decoder and introduce adapters between transformer blocks. 
Subsequently, we optimize this variant using the SSUL framework on the PASCAL VOC 15-1 benchmark. 
As illustrated in Table~\ref{table:pre_vanilla_vit}, the ViT-B variant with adapters significantly achieves 53.40\% mIoU for novel classes, indicating substantial improvement in generalization even with a frozen feature extractor $f$ post base classes training. 
Surprisingly, additional adapters also benefit base classes, resulting in a 0.7\% mIoU improvement. 
These findings solidly support our first hypothesis.

\textbf{Secondly}, instead of freezing feature extractor $f$ post base classes training, \emph{fully fine-tuning $f$ with a distillation term in subsequent training stages yields benefits for novel classes}. 
We adapt SSUL to enable continuous parameter updates for $f$ throughout training, optimizing the corresponding ViT-B with a linear head. Additionally, introducing feature distillation as a regularization term in this SSUL variant enhances anti-catastrophic forgetting. 
In the lower section of Table~\ref{table:pre_vanilla_vit}, compared to ViT-B-based SSUL with a linear head, the fully fine-tuned counterpart achieves a significant ($\sim$15.4\%) improvement in novel classes mIoU. 
With the added distillation loss, the mIoU for novel classes further increases to 62.67\%, maintaining anti-catastrophic forgetting ability. 
These findings affirm our second observation. Consequently, both the two modifications indicate that leveraging ViT-B with a simple linear head yields promising continual segmentation results, 
inspiring the design of our continual semantic segmentation framework, \emph{i.e.}, ConSept.
\begin{figure*}[t]
\begin{center}
\includegraphics[width=\textwidth]{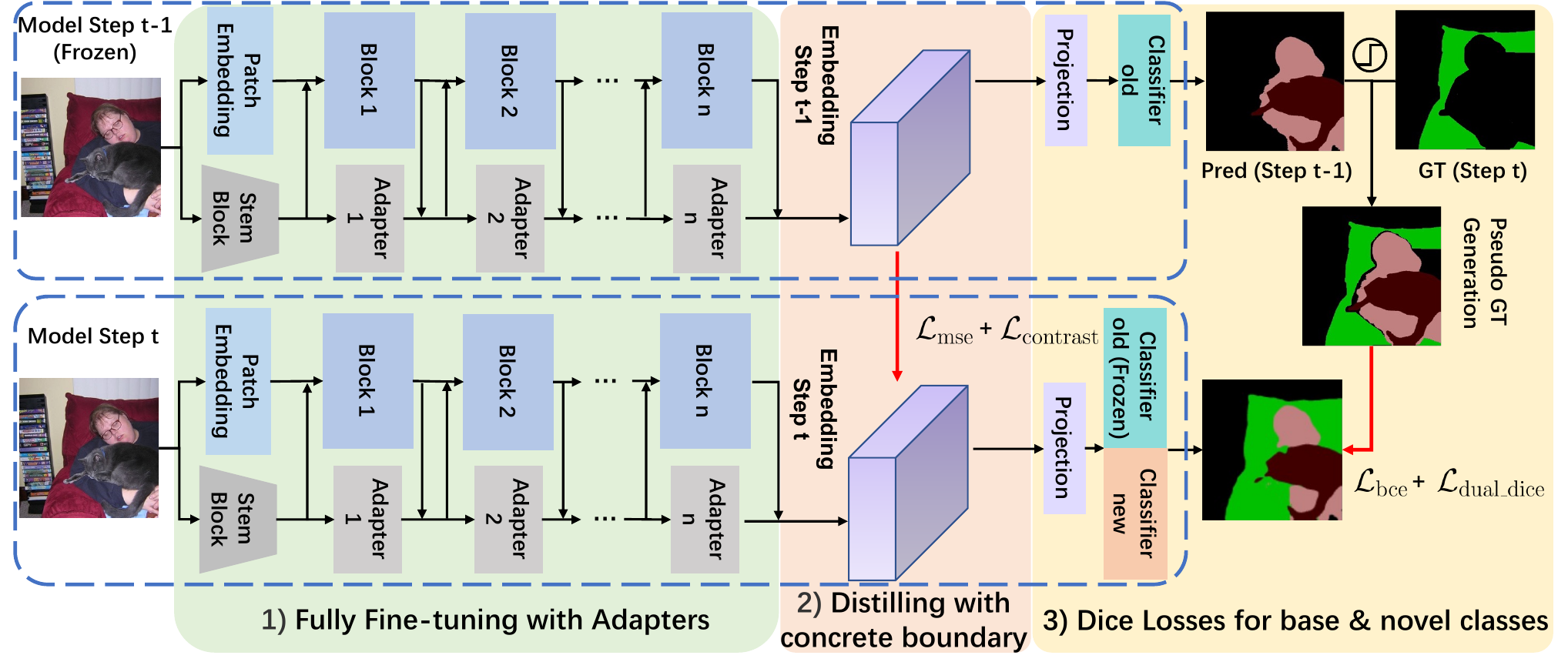}
\end{center}
\vspace{-2em}
\caption{
    Overview of our proposed ConSept. 
    The pipeline is primarily grounded on SSUL~\cite{cha2021ssul} by replacing the segmentation head with vanilla ViT accompanied with a linear head. 
    To fully harness the anti-catastrophic forgetting capability of ViT and enhance the generalization performance in continual segmentation scenarios, 
    we integrate adapters into ViTs, resulting in a dual-path feature extractor with a fully fine-tuning learning paradigm, which is the key element of ConSept. 
    Additionally, ConSept employs feature distillation with a frozen old-class linear head to enhance its anti-catastrophic forgetting ability and  incorporate dual dice losses to regularize the segmentation maps for overall segmentation performance.
}
\vspace{-1em}
\label{fig:pipeline}
\end{figure*}
\section{Proposed Method}\label{sec:method}
Building upon the insights gained in Sec.~\ref{sec:pre_analysis}, we propose to utilize the ImageNet-pretrained vanilla ViT feature extractor and a straightforward linear segmentation decoder to attain robust anti-catastrophic forgetting capability and generalization performance to novel classes within an appropriate training framework. 
In this section, we present our perspectives on training such a ViT-based continual semantic segmentation network and introduce the proposed solution, namely ConSept. 

\subsection{Overview of ConSept}
Fig.~\ref{fig:pipeline} delineates the comprehensive pipeline of our proposed ConSept.
In the training phase of step $t (t>1)$, we consider an input image $\mathbf{I}$ with the corresponding ground-truth segmentation mask $S_{t}$ randomly sampled from $\mathcal{D}_{t}$. 
The mask annotation $\mathbf{\hat{M}}$ exclusively encompasses classes in $\mathcal{C}_{t}$.
The input image $\mathbf{I}$ undergoes processing through the feature extractor $f_{t-1}$ optimized in the preceding $t-1$ learning steps. 
Here, $f_{t-1}$ encompasses both a vision transformer~\cite{dosovitskiy2020image} and the associated ViT-Adapter~\cite{chen2023vision}, yielding multi-scale features $\mathbf{F}_{t-1} = \{\mathbf{F}^{0}_{t-1}, ... ,\mathbf{F}^{3}_{t-1}\}$. 
For simplicity, we denote by $\mathbf{F}^{0}_{t-1}$ the feature from the shallow layers of $f_{t-1}$ with the largest spatial resolution, and $\mathbf{F}^{3}_{t-1}$ conversely.
Following the approach of \cite{xie2021segformer}, all features in $\mathbf{F}_{t-1}$ are interpolated to a consistent spatial resolution and concatenated into a fused feature $\mathbf{F}^{\text{fuse}}_{t-1}$. 
Subsequently, a linear projection layer $h_{t-1}$ maps $\mathbf{F}^{\text{fuse}}_{t-1}$ to $\mathbf{\hat{F}}^{\text{fuse}}_{t-1}$. 
Finally, the linear segmentation head $g_{1:t-1}$ of old classes is employed to predict segmentation masks $\mathbf{M}_{t-1}$ for classes in $\mathcal{C}_{1:t-1}$.
With the same architecture, the input image $\mathbf{I}$ is also processed through the $t$-step feature extractor $f_{t}$ with the corresponding projection layer $h_{t}$, yielding multi-scale features $\mathbf{F}_{t}$ and projected fused features $\mathbf{\hat{F}}^{\text{fuse}}_{t}$.
Then, to predict the segmentation masks for old and new categories, both the linear head of old classes $g_{1:t-1}$ and the counterpart of new classes $g_{t}$ are utilized as follows: 
\begin{equation}\label{eqn:step_t_segmap}
    \mathbf{M}_{t-1} = [g_{1:t-1}(\mathbf{\hat{F}}^{\text{fuse}}_{t}), g_{t}(\mathbf{\hat{F}}^{\text{fuse}}_{t})],
\end{equation}
where $[\dots]$ means the concatenate operation. 

During optimization, we first map the predicted segmentation mask $\mathbf{M}_{t-1}$ into the pseudo ground-truth mask $\mathbf{\hat{S}}_{1:t-1}$ as follows:
\begin{equation}\label{eqn:map}
    \mathbf{\hat{S}}_{1:t-1} = \max_{\mathcal{C}_{1:t-1}} \sigma (\mathbf{M}_{t-1}).
\end{equation}
Subsequently, the pseudo ground-truth mask $\mathbf{\hat{S}}_{1:t-1}$ of old classes is merged with the ground-truth mask $\mathbf{S}_{t}$ for new classes in $\mathcal{C}_{t}$ to form the ultimate pseudo ground-truth mask $\mathbf{\hat{S}}_{1:t}$ encompassing classes in $\mathcal{C}_{1:t}$. Specifically, for each pixel $i$, the pseudo ground-truth label $\mathbf{\hat{S}}_{1:t,i}$ is defined as follows:
\begin{equation}\label{eqn:merge_pseudo_gt}
    \mathbf{\hat{S}}_{1:t,i} = \left\{
    \begin{aligned}
        & \mathbf{S}_{t,i} & (\mathbf{S}_{t,i}\in \mathcal{C}_{t}) \\
        & \mathbf{\hat{S}}_{1:t-1,i} & (\mathbf{S}_{t,i}\notin \mathcal{C}_{t})
    \end{aligned}\right .
\end{equation}
Different from \cite{cha2021ssul,zhang2022microseg}, ConSept does not introduce the ``unknown'' class, but it still obtains robust and promising performance.
Finally, the segmentation loss $\mathcal{L}_{\text{bce}}+\mathcal{L}_{\text{dual\_dice}}$ between $\mathbf{M}_{t}$ and $\mathbf{\hat{S}}_{1:t}$ is minimized while minimizing the distillation loss $\mathcal{L}_{\text{mse}}+\mathcal{L}_{\text{contrast}}$ between $\mathbf{F}_{t}$ and $\mathbf{F}_{t-1}$. 

\subsection{Better Generalization: Fine-tuning with Adapters}
Given the notable progress of CNN-based visual adapters~\cite{nie2022protuning,chen2022convadapter,chen2023vision} in transfer learning, 
and in accordance with our first observation in Sec.~\ref{sec:pre_analysis}, 
one may hypothesize that introducing adapters or fine-tuning the feature extractor $f_{t}$ can enhance the generalization ability of novel classes, 
thereby improving the overall segmentation performance. 
Therefore, we propose to integrate lightweight adapters into ConSept. 
Specifically, ConSept comprises a shallow convolution-based stem block~\cite{xiao2021early} and multiple lightweight cross-attention layers to construct adapters. 
As there is no interaction between old and new models, we omit the step number $t$ in this context for simplicity. 
To elaborate, we initially utilize the stem block to extract multi-scale features, followed by flattening these features through a flatten operation, 
resulting in the initialized adapter feature $\mathbf{x}^{0}_{\text{ada}}$:
\begin{equation}
    \mathbf{x}^{0}_{\text{ada}} = \text{flatten}(\text{stem}(\mathbf{I})).
\end{equation}
Then for the $l$-th ViT-based feature $\mathbf{x}^{l}_{\text{vit}}$ generated by the $l$-th group of transformer blocks, we aggregate $\mathbf{x}^{l}_{\text{ada}}$ into $\mathbf{x}^{l}_{\text{vit}}$ by using cross-attention~\cite{vaswani2017attention}, 
\begin{equation}\label{eqn:adapter2vit}
    \mathbf{x}^{l}_{\text{vit}} = \mathbf{x}^{l}_{\text{vit}} + \text{Attn}(\text{norm}(\mathbf{x}^{l}_{\text{vit}}), \text{norm}(\mathbf{x}^{l}_{\text{ada}}), \text{norm}(\mathbf{x}^{l}_{\text{ada}})),
\end{equation}
where $\text{norm}(\cdot)$ means  LayerNorm~\cite{ba2016layer}, and $\text{Attn}(\textbf{q},\textbf{k},\textbf{v})$ means  cross-attention~\cite{vaswani2017attention}.
Subsequently, we refine the adapter feature using a feed-forward network and the cross-attention with the $(l+1)$-th ViT-based feature $\mathbf{x}^{l+1}_{\text{vit}}$, yielding $\mathbf{x}^{l+1}_{\text{ada}}$ as follows:
\begin{equation}\label{eqn:ffn}
    \begin{aligned}
        \mathbf{\hat{x}}^{l}_{\text{ada}} &= \mathbf{x}^{l}_{\text{ada}} + \text{Attn}(\text{norm}(\mathbf{x}^{l}_{\text{ada}}),\text{norm}(\mathbf{x}^{l+1}_{\text{vit}}),\text{norm}(\mathbf{x}^{l+1}_{\text{vit}})) \\
        \mathbf{x}^{l}_{\text{ada}} &= \mathbf{\hat{x}}^{l}_{\text{ada}} + \text{FFN}(\text{norm}(\mathbf{\hat{x}}^{l}_{\text{ada}}))
    \end{aligned},
\end{equation}
where $\text{FFN}$ means the feed-forward network. 
By feature aggregation with $n$ individual adapters, 
we acquire the ultimate adapter feature $\mathbf{x}^{n}_{\text{ada}}$, which is then split and reshaped to form $\mathbf{F}$.
For simplicity, in ConSept, we set $n=4$ as the number of adapters, 
striking a balance between performance and additional parameters.

The incorporation of adapters into ConSept offers dual advantages.
Firstly, the usage of lightweight and efficient modules (\emph{i.e.}, with fewer than 10\% additional parameters) in the feature extractor significantly enhances the generalization ability of ViT in class-incremental scenarios, \emph{without adversely affecting the optimization procedure or the final performance}.
Secondly, our proposed method remains compatible with other techniques that employ more complicated segmentation decoders~\cite{zhang2022microseg,shang2023incrementer,zhang2023coinseg} or additional supervision~\cite{zhang2022microseg,zhang2023coinseg} to achieve superior performance.

Moreover, conventional approaches~\cite{cha2021ssul,zhang2022microseg} typically freeze the parameters of the feature extractor during training steps with $t>1$, updating only the parameters of the segmentation decoder. 
This strategy aims to preserve the feature representation and segmentation ability of base classes.
In contrast to these methods, in the training phases for base and novel classes, we concurrently update parameters for both the feature extractor $f_{t}$ and the segmentation decoder $g_{t}$. 
Our approach allows $f_{t}$ to accumulate both texture and semantic information related to novel classes $\mathcal{C}_{t}$, ensuring ConSept to correctly localize regions from $\mathcal{C}_{t}$ and avoid suboptimal segmentation performance on $\mathcal{C}_{t}$. 
The experimental results in Sec.~\ref{sec:ablation} will verify the impact of incorporating adapters for ViT and adopting full fine-tuning in ConSept.
%

In addition to the pivotal macro design of ConSept, we introduce two additional strategies in the subsequent sections to enhance the anti-catastrophic forgetting ability and overall segmentation performance of ConSept.

\subsection{Better Anti-Forgetting: Distillation with Deterministic Old-classes Boundary}
As shown in Table~\ref{table:pre_vanilla_vit}, simple fine-tuning will inevitably lead to forgetting in ConSept, particularly for $\mathcal{C}_{1:t-1}$. 
We address this issue by introducing a feature distillation constraint on $\mathbf{F}_{t}$ through our proposed dual feature distillation loss $\mathcal{L}_{\text{distill}}$. 
The key components of $\mathcal{L}_{\text{distill}}$ include two aspects. 
Firstly, to ensure the consistency between features from the $t$-step network $\mathbf{F}_{t}$ and their corresponding old features $\mathbf{F}_{t-1}$, we employ the dense mean-square loss $\mathcal{L}_{\text{mse}}$ as the primary distillation loss. 
Specifically, for each $\mathbf{F}^{i}_{t}$, it is defined as follows:
\begin{equation}\label{eqn:l_mse}
    \mathcal{L}_{\text{mse}} = \sum_{i} (\mathbf{F}^{i}_{t} - \mathbf{F}^{i}_{t-1})^2.
\end{equation}

Secondly, vanilla $\mathcal{L}_{\text{mse}}$ may incorrectly remain regions from novel classes as the same as $\mathbf{F}_{t-1}$, 
potentially limiting the performance on $\mathcal{C}_{t}$. To enhance the discriminative ability of features from different regions, 
we introduce the contrastive distillation loss $\mathcal{L}_{\text{contrast}}$ between $\mathbf{F}_{t}$ and $\mathbf{F}_{t-1}$ as follows:
\begin{equation}
    \mathcal{L}_{\text{contrast}} = \sum_{i} \sum_{j} -\log\frac{\langle \mathbf{F}^{i}_{t,j}, \mathbf{F}^{i}_{t-1,j}\rangle}{\sum_{k} \langle \mathbf{F}^{i}_{t,j}, \mathbf{F}^{i}_{t-1,k}\rangle},
\end{equation}
where $\langle\cdot,\cdot\rangle$ means cosine similarity, $i$ means features from the $i$-th layer, $j$ and $k$ means the indices of pixels. 
Therefore, our proposed dual distillation loss $\mathcal{L}_{\text{distill}}$ is formulated as follows:
\begin{equation}
    \mathcal{L}_{\text{distill}} = \mathcal{L}_{\text{mse}} + \mathcal{L}_{\text{contrast}}.
\end{equation}
During experiments, we exclusively employ $\mathcal{L}_{\text{distill}}$ on feature $\mathbf{F}^{3}_{t}$ to prevent overfitting. The rationale and impact of this specific choice will be discussed in Section~\ref{sec:ablation}.

Moreover, we introduce anti-forgetting constraints into the lightweight segmentation decoder to further maintain performance on base classes. 
Assuming that during step $t-1$, ConSept has achieved optimal segmentation performance on $\mathcal{C}_{1:t-1}$, then the learned linear segmentation head represents an ideal decision boundary for $\mathcal{C}_{1:t-1}$. We utilize this decision boundary to constrain $f_{t}$. 
Consequently, we fine-tune the old classes with only a frozen linear head. 
Specifically, during training for base classes, we update all ConSept parameters to ensure optimal segmentation ability. 
For $t>1$, we freeze the parameters of the linear head for old classes $g_{1:t-1}$ and keep the other parameters updated during training. 
Then the predictions $\mathbf{M}_{t}$ from ConSept in step $t$ will provide confidence scores for old classes in $\mathbf{M}_{t}$ constrained by a fixed segmentation boundary. 
This restriction ensures that the corresponding feature $\mathbf{F}_{t}$ exhibits little or no variation in the region of base classes, thereby enhancing the anti-catastrophic forgetting ability. 
The effect of the distillation loss and frozen head will be revealed in the experimental results in Section~\ref{sec:ablation}. 
We will discuss how to further enhance the continual segmentation performance of ConSept through additional regularization in the following sections.

\subsection{Better Regularization: Dual Dice Losses for Segmentation}
We further enhance the training of ConSept in terms of segmentation loss. While the binary cross-entropy loss with fused pseudo ground-truth masks offers dense supervision for segmentation predictions, two issues remain in the context of continual segmentation tasks. First, the highly imbalanced distribution of pseudo or ground-truth masks between old and new classes constrain the performance of old classes. Second, the limited training data and iterations in continual learning steps hamper the discriminative ability for new classes. Therefore, the introduction of additional regularization for segmentation predictions is crucial for the efficacy of ConSept.

With this motivation, we propose dual dice losses $\mathcal{L}_{\text{dual\_dice}}$ for ConSept. $\mathcal{L}_{\text{dual\_dice}}$ comprises two integral components, namely, class-specific dice loss $\mathcal{L}_{\text{c-dice}}$ and old-new dice loss $\mathcal{L}_{\text{on-dice}}$. Specifically, to ensure that old classes are well learned and the catastrophic forgetting for new classes can be reduced, we compute the multi-class dice loss~\cite{milletari2016v,sudre2017generalised} between the predictions $\mathbf{M}_{t}$ in step $t$ and the corresponding pseudo ground-truths $\mathbf{\hat{S}}_{1:t}$, as shown below:
\begin{equation}\label{eqn:c-dice}
    \mathcal{L}_{\text{c-dice}} = \frac{1}{|\mathcal{C}_{1:t}|}\sum_{c\in \mathcal{C}_{1:t}}\frac{2\sum_{j=1}^{N}{\mathbf{M}[c]}_{t,j}{\mathbf{\hat{S}}[c]}_{1:t,j}}{\sum_{i=1}^{N}{\mathbf{M}[c]}_{t,j}^{2} + \sum_{i=1}^{N}{{\mathbf{\hat{S}}[c]}_{1:t,j}^{2}}}
\end{equation}
where $[c]$ indicates the confidence score or one-hot ground-truth label for class $c$ in $\mathbf{M}_{t}$ or $\mathbf{\hat{S}}_{1:t}$, $j$ and $N$ indicate the index as well as the total number of pixel, respectively. 

The old-new dice loss is to enhance the discriminative ability for novel classes in step $t$. To achieve this goal, an intuitive approach is to binarize the multi-class predictions along with the corresponding pseudo masks into binary counterparts, \emph{i.e.}, ``old'' class and ``new'' class. Specifically, for $\mathbf{M}_{t}$, we aggregate confidence scores of classes in $\mathcal{C}_{1:t-1}$ to obtain the confidence score of the ``old'' class and aggregate confidence scores of classes in $\mathcal{C}_{t}$ to obtain the score of the ``new'' class. For $\mathbf{\hat{S}}_{1:t}$, we first convert it into the corresponding one-hot label formulation and obtain $\mathbf{\hat{S}}^{0\text{-}1}_{1:t}$; then, we binarize $\mathbf{\hat{S}}^{0\text{-}1}_{1:t}$ into old classes and new classes. The overall conversion procedure is summarized as:
\begin{equation}\label{eqn:binarize}
    \begin{aligned}
        \mathbf{\tilde{M}}_{t} &= [\sum(\mathbf{M}_{t}[\mathcal{C}_{1:t-1}]), \sum(\mathbf{M}_{t}[\mathcal{C}_{t}])] \\
        \mathbf{\tilde{S}}_{1:t} &= [\max(\mathbf{\hat{S}}^{0\text{-}1}_{1:t}[\mathcal{C}_{1:t-1}]), \max(\mathbf{\hat{S}}^{0\text{-}1}_{1:t}[\mathcal{C}_{t}])]
    \end{aligned}
\end{equation}
Finally, we apply dice loss for $\mathbf{\tilde{M}}_{t}$ and $\mathbf{\tilde{S}}_{1:t}$ as follows:
\begin{equation}\label{eqn:on-dice}
    \mathcal{L}_{\text{on-dice}} = \frac{2\sum_{j=1}^{N}\mathbf{\tilde{M}}_{t,j}\mathbf{\tilde{S}}_{1:t,j}}{\sum_{j=1}^{N}\mathbf{\tilde{M}}_{t,j}^{2} + \sum_{j=1}^{N}{\mathbf{\tilde{S}}_{1:t,j}^{2}}}
\end{equation}
where $j$ and $N$ indicate the index as well as the total number of pixel, respectively. 

Finally, we summarize the training loss $\mathcal{L}_{t} (1\leq t \leq T)$ for optimizing ConSept. During the training of step 1 (\emph{i.e.}, training on base classes), only training data with annotations for base classes are involved in the optimization. The distillation loss $\mathcal{L}_{\text{distill}}$ is not calculated. Only the segmentation loss (\emph{i.e.}, binary cross-entropy loss $\mathcal{L}_{\text{bce}}$ and class-specific dice loss~\cite{milletari2016v} $\mathcal{L}_{\text{c-dice}}$) between the prediction $\mathbf{M}_{1}$ and the corresponding ground-truth mask $\mathbf{S}_{1}$ is considered:
\begin{equation}\label{eqn:l_1}
    \mathcal{L}_{t\text{=}1} = \mathcal{L}_{\text{bce}} + \mathcal{L}_{\text{c-dice}}.
\end{equation}
For training at any step $t$ with $t>1$ (\emph{i.e.}, training on tasks with novel classes), both the distillation loss $\mathcal{L}_{\text{distill}}$ and the dual dice losses $\mathcal{L}_{\text{dual\_dice}}$ are considered. Therefore, the training objective at step $t$ ($t>1$) is expressed as follows:
\begin{equation}
    \mathcal{L}_{t>1} = \mathcal{L}_{\text{bce}} + \mathcal{L}_{\text{c-dice}} + \mathcal{L}_{\text{on-dice}} + \mathcal{L}_{\text{mse}} + \mathcal{L}_{\text{contrast}}.
\end{equation}
\section{Experiments}
\subsection{Datasets and Metrics}
\textbf{Datasets}. 
Aligned with prior state-of-the-art continual semantic segmentation methods~\cite{zhang2022microseg,cermelli2020modeling,shang2023incrementer}, 
we assess ConSept on the PASCAL VOC dataset~\cite{everingham2010pascal} and ADE20K dataset~\cite{zhou2017scene}. 
The PASCAL VOC dataset comprises 10,582 training images and 1,449 validation images, featuring 20 foreground classes and 1 background class. 
The ADE20K dataset encompasses approximately 20,000 training images and 2,000 validation images, presenting a more challenging scenario with 150 categories compared to PASCAL VOC.

\textbf{Protocols}.
As in prior research~\cite{zhang2022microseg,cermelli2020modeling,shang2023incrementer}, 
on the PASCAL VOC dataset, we assess ConSept using 15-1, 15-5, and 19-1 tasks in both \emph{overlapped} and \emph{disjoint} settings. 
Task ``X-Y'' denotes using the initial $X$ categories as base classes and incorporating $Y$ novel classes for each new task. 
On the ADE20K dataset, we evaluate ConSept using 100-10, 100-50, and 50-50 tasks under the \emph{overlapped} setting.

\textbf{Metrics}.
In alignment with prior methods~\cite{zhang2022microseg,cermelli2020modeling,shang2023incrementer}, we employ the mean Intersection-over-Union (mIoU) as the primary metric for performance comparison with state-of-the-art methods.

\subsection{Implementation Details}
Building upon prior works in continual semantic segmentation for both CNNs~\cite{zhang2022microseg,cermelli2020modeling,douillard2021plop,michieli2021continual} and ViTs~\cite{cermelli2023comformer,shang2023incrementer}, 
ConSept is initialized with ImageNet-pretrained ViT-B parameters~\cite{dosovitskiy2020image,deng2009imagenet}. 
The newly added convolution layers employ Kaiming normal distribution~\cite{He2016DeepRL}, and linear layers use truncated normal distribution~\cite{burkardt2014truncated,dosovitskiy2020image} for weight initialization. 
The input shape is $512\times 512$, and fundamental data augmentation techniques (\emph{i.e.}, random crop and resize, and random horizontal flipping) are used in training and evaluation. 
During training, we leverage the AdamW optimizer~\cite{loshchilov2018decoupled} with a weight decay of 0.01 to optimize ConSept. 
The base task training follows previous ViT-based segmentation methods~\cite{cheng2021mask2former,strudel2021segmenter,xie2021segformer} for 100 epochs, while we halve the training epochs of novel tasks to mitigate catastrophic forgetting. 
The initial learning rate is $2\times 10^{-5}$, and the learning rate is multiplied by 10 for linear segmentation head. 
Following~\cite{chen2017deeplab}, we  leverage the polynomial learning rate scheduler with a power of $0.9$ to adjust learning rate during training. 
All codes are implemented by PyTorch toolkit~\cite{NEURIPS2019_9015}, and all experiments are
conducted on two NVIDIA RTX A6000 GPUs.

\begin{table*}[t]
\centering
    \caption{Comparison of diverse continual semantic segmentation methods on PASCAL VOC benchmarks under the \emph{overlapped} setting. Best results are \textbf{bolded}, and the second-best results are \underline{underlined}. ConSept achieves competitive performance across all the three tasks compared to state-of-the-art methods.}
    \label{table:compvoc_overlapped}
    \small
    \setlength{\tabcolsep}{7.0pt} 
   \renewcommand{\arraystretch}{4.0}
 { \fontsize{8.3}{3}\selectfont{
    \begin{threeparttable}
    \begin{tabular}{lcc|ccc|ccc|ccc}
    \bottomrule
    \multirow{2}{*}{\textbf{Method}}&\multirow{2}{*}{\textbf{Backbone}}&\multirow{2}{*}{\textbf{Decoder}}&\multicolumn{3}{c}{\textbf{15-1 (6 steps)}}&\multicolumn{3}{c}{\textbf{15-5 (2 steps)}}&\multicolumn{3}{c}{\textbf{19-1 (2 steps)}}
    \\ \cline{4-12}
    &&&{{0-15}}&{{16-20}}&{all}&{{0-15}}&{{16-20}}&{all}&{{0-19}}&{{20}}&{all}
    \\ \hline
    \multicolumn{7}{l}{ \textbf{CNN-based Methods}  }\\
    \hline
    ILT~\cite{michieli2019incremental} & ResNet-101 & DeepLab V3 & 8.75 & 7.99 & 8.56 & 67.48 & 39.23 & 60.45 & 67.75 & 10.88 & 65.05  \\
    MiB~\cite{cermelli2020modeling} & ResNet-101 & DeepLab V3 & 34.22 & 13.50 & 29.29 & 76.37 & 49.97 & 70.08 & 71.43 & 23.59 & 69.15 \\
    SDR~\cite{michieli2021continual} & ResNet-101 & DeepLab V3 & 44.70 & 21.80 & 39.20 & 75.40 & 52.60 & 69.90 & 69.10 & 32.60 & 67.40 \\
    PLOP~\cite{douillard2021plop} & ResNet-101 & DeepLab V3 & 65.12 & 21.11 & 54.64 & 75.73 & 51.71 & 70.09 & 75.35 & 37.35 & 73.54\\
    RECALL~\cite{maracani2021recall} & ResNet-101 & DeepLab V3 & 65.70 & 47.80 & 62.70 & 66.60 & 50.90 & 64.00 & 67.90 & 53.50 & 68.40 \\
    REMIND~\cite{phan2022class} & ResNet-101 & DeepLab V3 & 68.30 & 27.23 & 58.52 & 76.11 & 50.74 & 70.07 & 76.48 & 32.34 & 74.38 \\
    SSUL~\cite{cha2021ssul} & ResNet-101 & DeepLab V3 & 78.40 & 49.00 & 71.40 & 78.40 & 55.80 & 73.00 & 77.80 & 49.80 & 76.50 \\
    {MicroSeg}~\cite{zhang2022microseg}&{ResNet-101}& MicroSeg & 81.30 &{52.50}&{74.40} & 82.00 & 59.20 & 76.60 & 79.30 & 62.90 & 78.50 \\
    \hline
    {Joint (CNN)} \textbf{Upper Bound} & ResNet-101 & DeepLab V3 & 82.70 & 75.00 & 80.90 & 82.70 & 75.00 & 80.90 & 81.0 & 79.1 & 80.9  \\
    \toprule
    \multicolumn{7}{l}{ \textbf{ViT-based Methods}  }\\
    \hline
    {MiB}~\cite{cermelli2020modeling}&{ViT-Base}& DeepLab V3 & {72.55} &{23.14}&{61.73}&{78.62}&{63.10}&{75.62}&{79.91}&{47.70}&{79.10} \\
    {RBC}~\cite{zhao2022rbc}&{ViT-Base}& DeepLab V3 &{75.90}&{40.15}&{68.24}&{78.86}&{62.01}&{75.53}&{80.24}&{38.79}&{78.99} \\
    {MicroSeg}~\cite{zhang2022microseg}&{Swin-Base}&MicroSeg &{82.00}&{47.30}&{73.70}&{82.90}&{60.10}&{77.50}&{81.00}&{62.40}&{80.00} \\
    {CoinSeg}~\cite{zhang2023coinseg}&{Swin-Base}&MicroSeg&\underline{84.10}&\underline{65.60}&\underline{79.60}&\underline{84.10} & \underline{69.90} & \underline{80.80} & \underline{82.70} & {52.60} & {79.80} \\
    {Incrementer}~\cite{shang2023incrementer}&{ViT-Base}& Segmenter &{79.60}&{59.56}&{75.55}&{82.53}&{69.25}&{79.93}&{82.54}&\underline{60.95}&\underline{82.14} \\
    \textbf{ConSept (Ours)}&{ViT-Base}& Linear  &\textbf{84.53}&\textbf{66.06}&\textbf{80.13}&\textbf{84.94}&\textbf{73.69}&\textbf{82.26}&\textbf{84.03}&\textbf{72.02}&\textbf{83.46} \\
    \hline
    Joint (ViT) \textbf{Upper Bound}& ViT-Base & Linear  & 85.66 & 82.69 & 84.95 & 85.66 & 82.69 & 84.95 & 84.97 & 84.63 & 84.95 \\
    \toprule
    \end{tabular}
    \end{threeparttable}}}
    \vspace{-1em}
\end{table*}

\begin{table*}[t]
  \centering
      \caption{Comparison of diverse continual semantic segmentation methods on PASCAL VOC benchmarks under the \emph{disjoint} setting. Best results are \textbf{bolded}, and the second-best results are \underline{underlined}. ConSept achieves competitive performance across all the three tasks compared to state-of-the-art methods.}
      \label{table:compvoc_disjoint}
      \small
      \setlength{\tabcolsep}{7.0pt} 
     \renewcommand{\arraystretch}{4.0}
   { \fontsize{8.3}{3}\selectfont{
      \begin{threeparttable}
      \begin{tabular}{lcc|ccc|ccc|ccc}
      \bottomrule
      \multirow{2}{*}{\textbf{Method}}&\multirow{2}{*}{\textbf{Backbone}}&\multirow{2}{*}{\textbf{Decoder}}&\multicolumn{3}{c}{\textbf{15-1 (6 steps)}}&\multicolumn{3}{c}{\textbf{15-5 (2 steps)}}&\multicolumn{3}{c}{\textbf{19-1 (2 steps)}}
      \\ \cline{4-12}
      &&&{{0-15}}&{{16-20}}&{all}&{{0-15}}&{{16-20}}&{all}&{{0-19}}&{{20}}&{all}
      \\ \hline
      \multicolumn{7}{l}{ \textbf{CNN-based Methods}  }\\
      \hline
      ILT~\cite{michieli2019incremental} & ResNet-101 & DeepLab V3 & 3.70 & 5.70 & 4.20 & 63.20 & 39.50 & 57.30 & 69.10 & 16.40 & 66.40  \\
      MiB~\cite{cermelli2020modeling} & ResNet-101 & DeepLab V3 & 46.20 & 12.90 & 37.90 & 71.80 & 43.30 & 64.70 & 69.60 & 25.60 & 67.40 \\
      SDR~\cite{michieli2021continual} & ResNet-101 & DeepLab V3 & 59.20 & 12.90 & 48.10 & 73.50 & 47.30 & 67.20 & 69.90 & 37.30 & 68.40 \\
      PLOP~\cite{douillard2021plop} & ResNet-101 & DeepLab V3 & 57.86 & 13.67 & 46.48 & 71.00 & 42.82 & 64.29 & 75.37 & 38.89 & 73.64\\
      RECALL~\cite{maracani2021recall} & ResNet-101 & DeepLab V3 & 66.00 & 44.90 & 62.10 & 66.30 & 49.80 & 63.50 & 65.20 & 50.10 & 65.80 \\
      \hline
      {Joint (CNN)} \textbf{Upper Bound} & ResNet-101 & DeepLab V3 & 82.70 & 75.00 & 80.90 & 82.70 & 75.00 & 80.90 & 81.0 & 79.1 & 80.9  \\
      \toprule
      \multicolumn{7}{l}{ \textbf{ViT-based Methods}  }\\
      \hline
      {MiB}~\cite{cermelli2020modeling}&{ViT-Base}& DeepLab V3 & {66.74} &{26.32}&{58.28}&{74.98}&{59.90}&{72.27}&{80.61}&{45.17}&{79.61} \\
      {RBC}~\cite{zhao2022rbc}&{ViT-Base}& DeepLab V3 &{69.03}&{28.37}&{60.54}&{77.70}&{59.06}&{74.05}&{80.94}&{42.05}&{79.68} \\
      {Incrementer}~\cite{shang2023incrementer}&{ViT-Base}& Segmenter &\textbf{81.42}&\underline{57.05}&\underline{76.25}&\underline{81.59}&\underline{62.17}&\underline{77.60}&\underline{82.39}&\underline{64.18}&\underline{82.15} \\
      \textbf{ConSept (Ours)}&{ViT-Base}& Linear  &\underline{80.54}&\textbf{63.88}&\textbf{76.57}&\textbf{82.06}&\textbf{71.79}&\textbf{79.62}&\textbf{83.92}&\textbf{71.31}&\textbf{83.32} \\
      \hline
      Joint (ViT) \textbf{Upper Bound}& ViT-Base & Linear  & 85.66 & 82.69 & 84.95 & 85.66 & 82.69 & 84.95 & 84.97 & 84.63 & 84.95 \\
      \toprule
      \end{tabular}
      \end{threeparttable}}}
      \vspace{-1em}
  \end{table*}
\subsection{Comparison with State-of-the-Art Methods}
\textbf{Results on PASCAL VOC}.
We evaluate the competing methods on PASCAL VOC benchmarks~\cite{everingham2010pascal} with the \emph{overlapped} setting in Table~\ref{table:compade_overlapped}. ConSept achieves state-of-the-art mIoU across all classes. 
Notably, without introducing intricate segmentation decoders~\cite{strudel2021segmenter}, 
ConSept outperforms Incrementer~\cite{shang2023incrementer} by 4.58\%, 2.33\%, and 1.32\% in all-classes mIoU for 15-1, 15-5, and 19-1 tasks, respectively. 
Compared to CoinSeg~\cite{zhang2023coinseg}, which leverages a superior pretrained feature extractor (\emph{i.e.}, ImageNet-pretrained~\cite{deng2009imagenet} Swin-Base~\cite{Liu_2021_ICCV}) and external foundation models' region proposal knowledge, 
ConSept still surpasses it by 0.53\%, 1.46\%, and 3.66\% in all-classes mIoU for 15-1, 15-5, and 19-1 tasks, respectively.
Furthermore, to assess our method's anti-catastrophic forgetting capability, 
we utilize the architecture of ConSept and retrain the oracle model for all categories as an upper bound of mIoU, 
depicted in the last row of Table~\ref{table:compvoc_overlapped}. 
ConSept exhibits approximately 1\% forgetting on base classes in mIoU, largely outperforming other methods while achieving superior mIoU on novel classes. 
These encouraging results affirm the effectiveness of ConSept.

In the more challenging \emph{disjoint} setting on PASCAL VOC~\cite{everingham2010pascal}, competing results are also obtained. As shown in Table~\ref{table:compvoc_disjoint}, ConSept shows state-of-the-art performance across all the three benchmarks. Specifically, ConSept achieves 79.62\% and 83.32\% mIoU for 15-5 and 19-1 tasks, respectively, surpassing Incrementer~\cite{shang2023incrementer} by 2.02\% and 1.17\%.

\begin{table*}[t]
  \centering
      \caption{Comparison of different continual semantic segmentation methods on ADE20K benchmarks with \emph{overlapped} setting. Best results are \textbf{bolded} and the second best results are \underline{underlined}. ConSept obtains competing performance against state-of-the-art methods on all the three tasks.}
      \label{table:compade_overlapped}
      \small
      \setlength{\tabcolsep}{7.0pt} 
     \renewcommand{\arraystretch}{4.0}
   { \fontsize{8.3}{3}\selectfont{
      \begin{threeparttable}
      \begin{tabular}{lcc|ccc|ccc|ccc}
      \bottomrule
      \multirow{2}{*}{\textbf{Method}}&\multirow{2}{*}{\textbf{Backbone}}&\multirow{2}{*}{\textbf{Decoder}}&\multicolumn{3}{c}{\textbf{100-50 (2 steps)}}&\multicolumn{3}{c}{\textbf{50-50 (3 steps)}}&\multicolumn{3}{c}{\textbf{100-10 (6 steps)}}
      \\ \cline{4-12}
      &&&{{0-15}}&{{16-20}}&{all}&{{0-15}}&{{16-20}}&{all}&{{0-19}}&{{20}}&{all}
      \\ \hline
      \multicolumn{7}{l}{\textbf{CNN-based Methods}  }\\
      \hline
      ILT~\cite{michieli2019incremental} & ResNet-101 & DeepLab V3 & 18.30 & 14.40 & 17.00 & 3.50 & 12.90 & 9.70 & 0.10 & 3.10 & 1.10  \\
      MiB~\cite{cermelli2020modeling} & ResNet-101 & DeepLab V3 & 40.52 & 17.17 & 32.79 & 45.57 & 21.01 & 29.31 & 38.21 & 11.12 & 29.24 \\
      SDR~\cite{michieli2021continual} & ResNet-101 & DeepLab V3 & 37.40 & 24.80 & 33.20 & 40.90 & 23.80 & 29.50 & 28.90 & 7.40 & 21.70 \\
      PLOP~\cite{douillard2021plop} & ResNet-101 & DeepLab V3 & 41.66 & 15.42 & 32.97 & 47.75 & 21.60 & 30.43 & 39.42 & 13.63 & 30.88\\
      RBC~\cite{zhao2022rbc} & ResNet-101 & DeepLab V3 & 42.90 & 21.49 & 35.81 & 49.59 & 26.32 & 34.18 & 39.01 & 21.67 & 33.27 \\
      REMIND~\cite{phan2022class} & ResNet-101 & DeepLab V3 & 41.55 & 19.16 & 34.14 & 47.11 & 20.35 & 29.39 & 38.96 & 21.28 & 33.11 \\
      SSUL~\cite{cha2021ssul} & ResNet-101 & DeepLab V3 & 42.80 & 17.50 & 34.40 & 49.10 & 20.10 & 29.80 & 42.90 & 17.70 & 34.50 \\
      {MicroSeg}~\cite{zhang2022microseg}&{ResNet-101}& MicroSeg & 43.40 &{20.90}&{35.90} & 49.80 & 22.00 & 31.40 & 43.70 & 22.20 & 36.60 \\
      \hline
      {Joint (CNN)} \textbf{Upper Bound} & ResNet-101 & DeepLab V3 & 43.90 & 27.20 & 38.30 & 50.90 & 32.10 & 38.30 & 43.90 & 27.20 & 38.30  \\
      \toprule
      \multicolumn{7}{l}{\textbf{ViT-based Methods}  }\\
      \hline
      {MiB}~\cite{cermelli2020modeling}&{ViT-Base}& DeepLab V3 & {46.40} &{34.95}&{42.58}&{52.21}&{35.56}&{41.11}&{42.95}&{30.80}&{38.90} \\
      {MicroSeg}~\cite{zhang2022microseg}&{Swin-Base}&MicroSeg &{41.10}&{24.10}&{35.40}&{49.80}&{23.90}&{32.50}&{41.00}&{22.60}&{34.80} \\
      {CoinSeg}~\cite{zhang2023coinseg}&{Swin-Base}&MicroSeg&{41.60}&{26.70}&{36.60}&{49.00} & {28.90} & {35.60} & {42.10} & {24.50} & {36.20} \\
      {Incrementer}~\cite{shang2023incrementer}&{ViT-Base}& Segmenter &\underline{49.42}&\underline{35.62}&\underline{44.82}&\underline{56.15}&\underline{37.81}&\underline{43.92}&\underline{48.47}&\textbf{34.62}&\underline{43.85} \\
      \textbf{ConSept (Ours)}&{ViT-Base}& Linear  &\textbf{51.42}&\textbf{36.58}&\textbf{46.51}&\textbf{56.92}&\textbf{38.63}&\textbf{44.81}&\textbf{49.37}&\underline{33.00}&\textbf{43.95} \\
      \hline
      Joint (ViT) \textbf{Upper Bound}& ViT-Base & Linear  & 51.39 & 39.02 & 47.30 & 57.41 & 42.14 & 47.30 & 51.39 & 39.02 & 47.30 \\
      \toprule
      \end{tabular}
      \end{threeparttable}}}
      \vspace{-2em}
  \end{table*}

\begin{figure}[t]
  \begin{center}
  \includegraphics[width=0.48\textwidth]{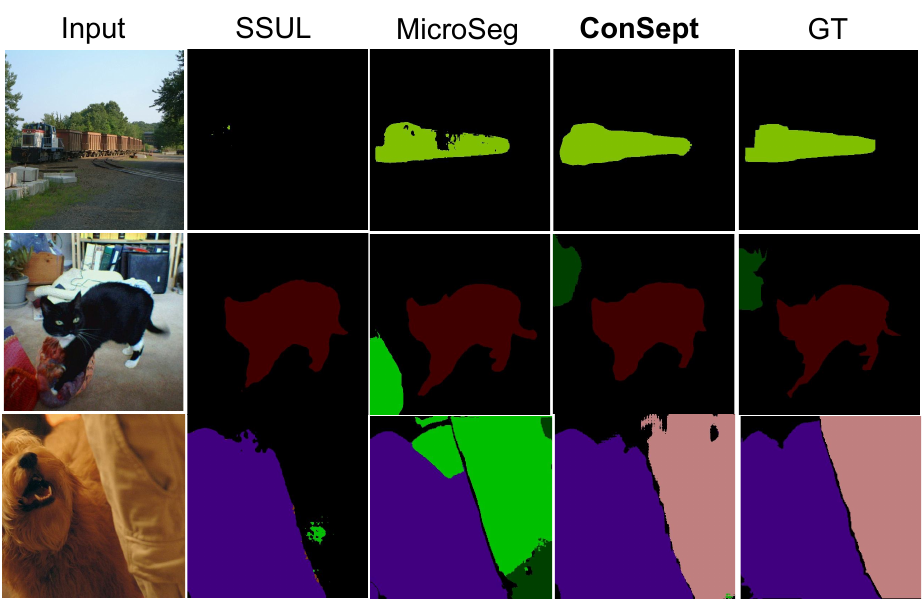}
  \end{center}
  \vspace{-1em}
  \caption{
      Visual comparison between previous state-of-the-art methods (\emph{i.e.}, SSUL~\cite{cha2021ssul}, MicroSeg~\cite{zhang2022microseg}) and ConSept on PASCAL VOC~\cite{everingham2010pascal} 15-1 benchmark under the \emph{overlapped} setting. Our method performs the best on both base and novel classes.
  }
  \vspace{-2em}
  \label{fig:vis_voc_comparison}
  \end{figure}

\textbf{Results on ADE20K Benchmarks}.
Then we compare ConSept with prior state-of-the-art methods on the more challenging ADE20K benchmarks. 
The results are shown in Table~\ref{table:compade_overlapped}. One can see that ConSept showcases competing performance with state-of-art-art methods. 
Specifically, ConSept attains 46.51\% and 44.81\% in mIoU for the 100-50 and 50-50 tasks, respectively, 
surpassing the state-of-the-art ViT-based method Incrementer~\cite{shang2023incrementer} by 1.7\% and 0.9\%. 
For the 100-10 task, ConSept achieves a competitive 43.95\% mIoU across all classes. 
Additionally, ConSept experiences less drop in mIoU on base classes across all the three tasks than Incrementer. 
These results demonstrate the effectiveness of ConSept on intricate and challenging scenarios.

\begin{table}[t]
  \centering
      \caption{Comparison of different continual semantic segmentation methods on PASCAL VOC 10-1 task with \emph{overlapped} setting. Best results are \textbf{bolded} and the second best results are \underline{underlined}. ConSept obtains competing performance against state-of-the-art methods. Meanwhile ConSept exhibits strong anti-catastrophic forgetting ability for both base classes and novel categories in different training steps.}
      \label{table:comp10-1}
      \small
      \setlength{\tabcolsep}{3.5pt} 
     \renewcommand{\arraystretch}{4.0}
   { \fontsize{8.3}{3}\selectfont{

      \begin{tabular}{l|ccc}
      \bottomrule
      \multirow{2}{*}{\textbf{Method}}&\multicolumn{3}{c}{\textbf{10-1 (11 steps)}}
      \\ \cline{2-4}
      &{{0-10}}&{{11-20}}&{all}
      \\ \hline
      Joint (CNN) &82.1&79.6&80.9 \\
      {MiB}~\cite{cermelli2020modeling}&20.0 (-62.1\%)&20.1 (-59.5\%)&20.1 (-60.8\%) \\
      SDR~\cite{michieli2021continual} &32.4 (-49.7\%)&17.1 (-62.5\%)&25.1 (-63.8\%) \\
      PLOP~\cite{douillard2021plop} &44.0 (-38.1\%)&15.5 (-64.1\%)&30.5 (-50.4\%) \\
      RECALL~\cite{maracani2021recall}&59.5 (-22.6\%)&46.7 (-32.9\%)&54.8 (-26.1\%) \\
      RCIL~\cite{zhang2022representation} &55.4 (-26.7\%)&15.1 (-64.5\%)&34.3 (-46.6\%) \\
      {SSUL (CNN)}~\cite{cha2021ssul}&74.0 (-8.1\%)&53.2 (-26.4\%)&64.1 (-16.8\%) \\
      {MicroSeg (CNN)}~\cite{zhang2022microseg}&77.2 (-4.9\%)&57.2 (-22.4\%)&67.7 (-13.2\%) \\
      \bottomrule
      Joint (Swin-B) &{82.4}&{83.0}&{82.7} \\
      {SSUL (Swin-B)}~\cite{cha2021ssul} &{75.3} {(-7.1\%)}&{54.1} {(-28.9\%)}&{65.2} {(-17.5\%)} \\
      {MicroSeg (Swin-B)}~\cite{zhang2022microseg}&\underline{78.9} \underline{\textcolor{blue}{(-3.5\%)}}&{59.2} {(-23.8\%)}&{70.1} {(-12.6\%)} \\
      {CoinSeg}~\cite{zhang2023coinseg}  & \textbf{81.3} \textbf{\textcolor{blue}{(-1.1\%)}} & \underline{64.4} \underline{\textcolor{blue}{(-18.6\%)}} & \textbf{73.7} \textbf{\textcolor{blue}{(-9.0\%)}} \\
      \hline
      Joint (ViT-B) & 84.4 & 85.5 & 84.9 \\
      {Incrementer}~\cite{shang2023incrementer}&{77.6} {(-6.8\%)}&{60.3} {(-25.2\%)}&{70.2} {(-14.6\%)} \\
      \textbf{ConSept (Ours)} &{77.5} {(-6.9\%)}&\textbf{69.5} \textbf{\textcolor{blue}{(-16.0\%)}}&\textbf{73.7} \underline{\textcolor{blue}{(-11.2\%)}} \\
      \toprule
      \end{tabular}
      }}
      \vspace{-1em}
  \end{table}

\textbf{Qualitative Results Analysis}.
In addition to quantitative results, we visualize the segmentation maps to compare ConSept with other methods. 
%
On the PASCAL VOC 15-1 benchmark~\cite{everingham2010pascal}, we compare SSUL~\cite{cha2021ssul}, MicroSeg~\cite{zhang2022microseg} and ConSept, all of which are based on the pretrained ViT,  under the \emph{overlapped} setting. 
As illustrated in Fig.~\ref{fig:vis_voc_comparison}, ConSept accurately localizes regions of base classes while generating more precise segmentation masks for novel classes. 
Specifically, ConSept produces more accurate masks for the novel ``train'' class and correctly localizes the newly-added ``\texttt{potted plant}'' and ``\texttt{person}'' classes.

\begin{figure*}[t]
\begin{center}
\includegraphics[width=\textwidth]{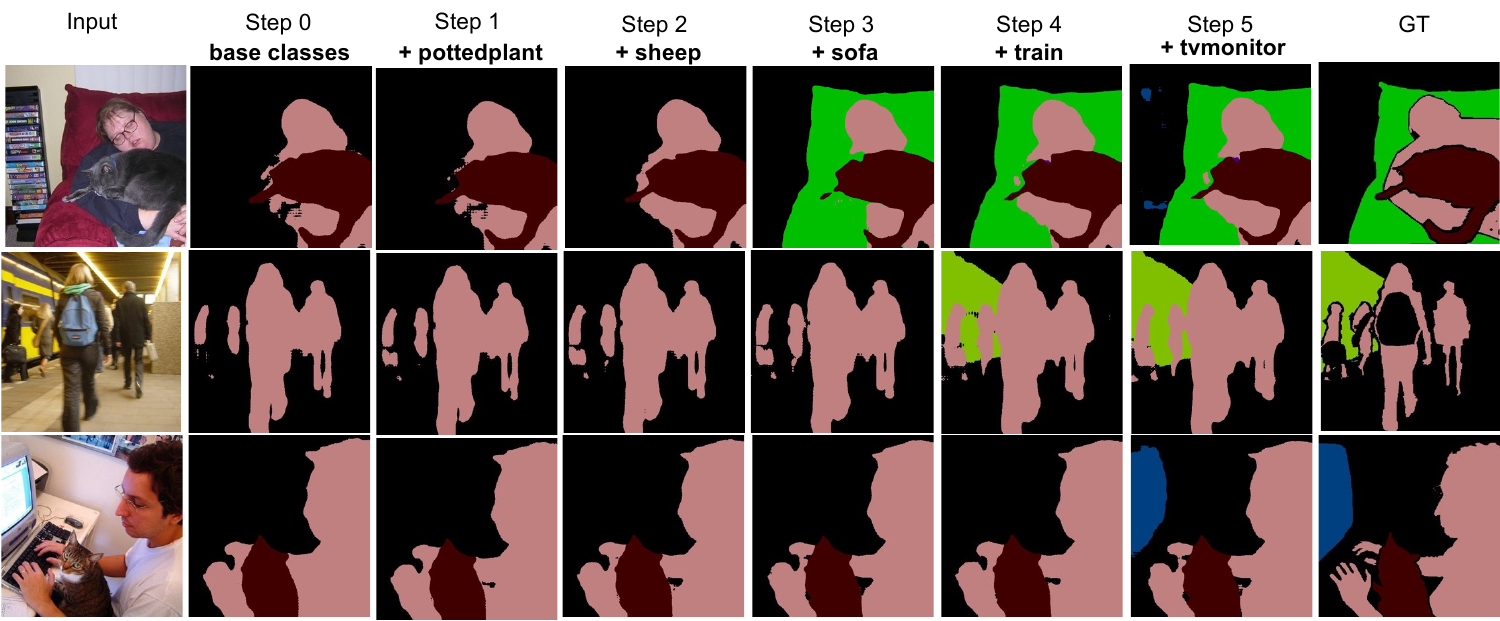}
\end{center}
\vspace{-1em}
\caption{
    Visualization of predictions from ConSept on PASCAL VOC 15-1 task with \emph{overlapped} setting. ConSept exhibits stable anti-catastrophic forgetting ability for old classes and good generalization ability for novel classes.
}
\vspace{-1.2em}
\label{fig:vis_per_step_15-1}
\end{figure*}

\begin{figure*}[t]
\begin{center}
\includegraphics[width=\textwidth]{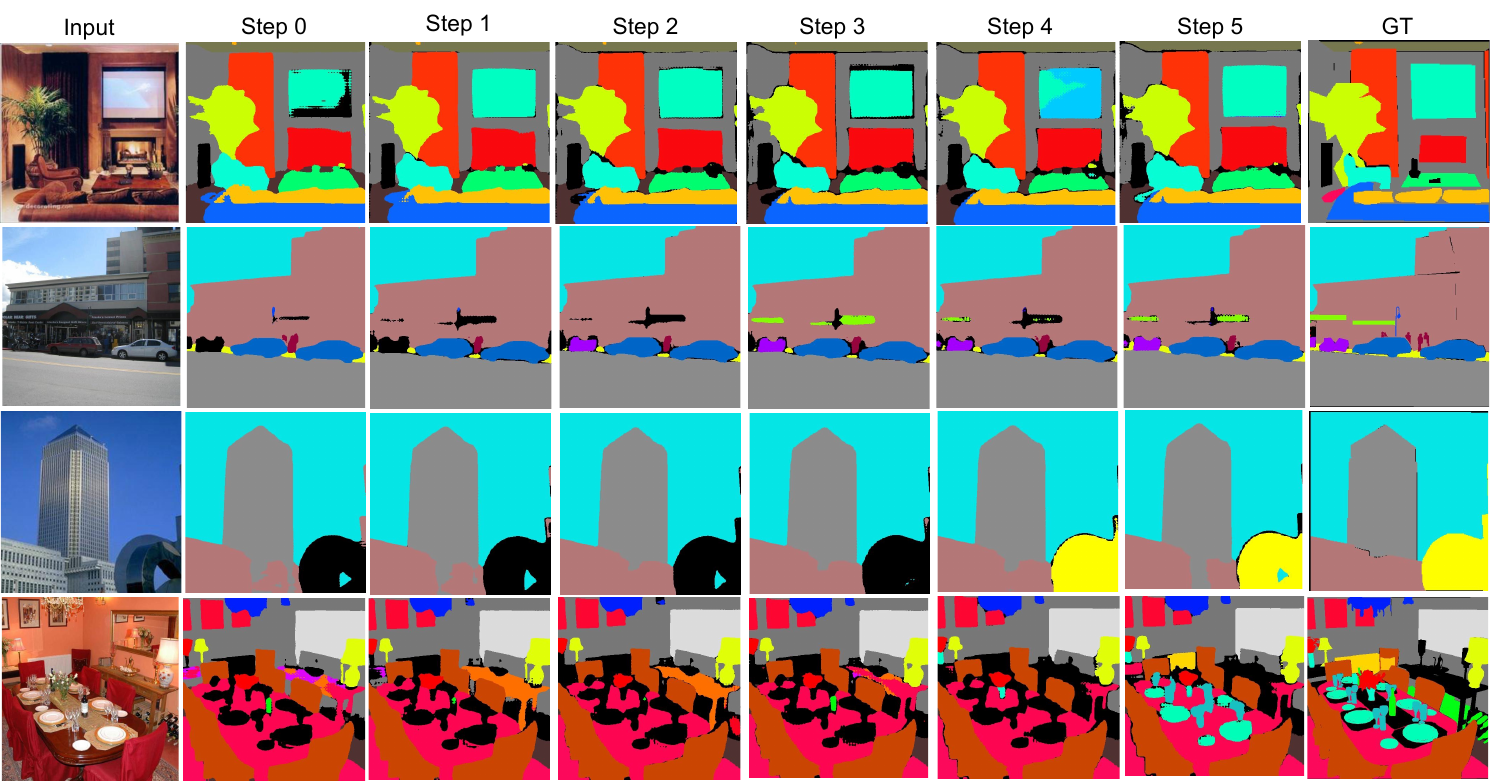}
\end{center}
\vspace{-1em}
\caption{
    Visualization of ConSept on ADE20K 100-10 task with \emph{overlapped} setting. ConSept performs well on more challenging tasks.
}
\vspace{-1.3em}
\label{fig:vis_per_step_100-10}
\end{figure*}

Furthermore, we explore whether ConSept can learn new concepts without forgetting segmentation capability for old classes. 
Fig.\ref{fig:vis_per_step_15-1} illustrates the per-step visualization results of ConSept on PASCAL VOC 15-1 benchmark\cite{everingham2010pascal}. 
ConSept successfully learns the concepts of ``\texttt{sofa}'', ``\texttt{train}'', and ``\texttt{tv monitor}'' after corresponding task learning, preserving segmentation quality on base classes. 
Additionally, Fig.\ref{fig:vis_per_step_100-10} depicts per-step visualization results of ConSept on ADE20K 100-10 benchmark\cite{zhou2017scene}. 
Even in more complex scenarios, ConSept maintains stable anti-catastrophic forgetting capability for old classes and strong generalization ability for novel classes. 

\textbf{Results on Longer Training Steps}.
To demonstrate the effectiveness of ConSept in tasks with extended training steps, we evaluate it on the PASCAL VOC 10-1 task (\emph{i.e.}, 11 steps in total) under the \emph{overlapped} setting. 
Given that the seen categories are introduced during the exceptionally long training steps, we evaluate the performance and anti-catastrophic forgetting ability of both old and new categories. 
Table~\ref{table:comp10-1} displays the results. ConSept exhibits state-of-the-art mIoU for all and novel classes, surpassing CoinSeg (Swin-B) by 5.1\% in mIoU for novel classes. 
Regarding the anti-catastrophic forgetting ability, compared to the corresponding joint training model, ConSept experiences only a 6.9\% and 11.2\% drop in mIoU for novel classes and all classes, respectively. 
Simultaneously, ConSept sustains less than 7\% decrease in mIoU for base classes, indicating a favorable trade-off between anti-catastrophic forgetting ability for base and novel classes. 
These promising results demonstrate the strong performance of ConSept in continual semantic segmentation with much more training steps.
  
\subsection{Ablation Studies and Analysis}\label{sec:ablation}
To examine each component of ConSept, we perform ablation studies and analyses on the PASCAL VOC 15-1 benchmark under the \emph{overlapped} setting in this section.

\textbf{Ablation Study}.
We systematically examine the impact of each essential component on ConSept in Table~\ref{table:ablation_study}. 
Type (a) is our baseline method, which employs ViT-B with a linear segmentation head as the continual segmenter and is optimized via the SSUL~\cite{cha2021ssul} framework. 
The baseline achieves 80.81\% base classes mIoU and 67.97\% overall mIoU, while reaches 26.87\% novel classes mIoU. 
Introducing adapters to vision transformers in type (b) results in a substantial performance improvement of 26.51\% in novel classes mIoU and 8.58\% in overall mIoU. 
Further enhancements in type (c) are achieved by applying full fine-tuning with a frozen old-classes linear head and regularization through distillation loss, 
leading to an additional $\sim$11\% improvement in novel classes mIoU while maintaining similar segmentation performance on base classes. 
These results suggest that ConSept benefits from fine-tuning with distillation to enhance generalization ability on new classes, while preserving sufficient anti-catastrophic forgetting capability on old classes. 
%
Finally, introducing the dual dice losses $\mathcal{L}_{\text{dual\_dice}}$ in the full ConSept yields an overall mIoU of 80.13\%. 
Notably, compared to full ConSept, the type (d) variant without $\mathcal{L}_{\text{distill}}$ 
experiences a degradation in overall mIoU to 77.65\%, with performance losses of $\sim$2.7\% and $\sim$1.6\% in terms of base and novel classes mIoU, respectively. 
In conclusion, the above evaluation results highlight the effectiveness of our proposed components.

\begin{table*}[t]
  \centering
      \caption{Ablation study of each key component in ConSept, where ``Linear'' means using linear segmentation head.
      }
      \vspace{-0.5em}
      \label{table:ablation_study}
      \small
      \setlength{\tabcolsep}{5.5pt} 
     \renewcommand{\arraystretch}{4.0}
   { \fontsize{8.3}{3}\selectfont{

      \begin{tabular}{lccccc|ccc}
      \bottomrule
      \multirow{2}{*}{\textbf{Type}}&\multirow{2}{*}{\textbf{Linear}}&\multirow{2}{*}{\textbf{Adapter}}&\multirow{2}{*}{\textbf{Tuning}}&\multirow{2}{*}{\bf $\mathcal{L}_{\text{distill}}$}&\multirow{2}{*}{\bf $\mathcal{L}_{\text{dual-dice}}$}&\multicolumn{3}{c}{\textbf{VOC 15-1 (6 steps)}}
      \\ \cline{7-9}
      &&&&&&{{0-15}}&{{16-20}}&{all}
      \\ \hline
      (a) & \checkmark & - & - & - & -  & 80.81 & 26.87 & 67.97 \\
      (a) & \checkmark & \checkmark & - & - & -  & 81.45 (+0.64\%) & 53.38 (+26.51\%) & 74.76 (+6.79\%) \\
      (c) & \checkmark & \checkmark & \checkmark & \checkmark & -  & 80.31 (-0.50\%)& 64.54 (+37.77\%) & 76.55 (+8.58\%) \\
      (d) & \checkmark & \checkmark & \checkmark & - & \checkmark &81.81 (+1.00\%)& 64.35 (+37.58\%) & 77.65 (+8.68\%) \\
      (e) & \checkmark& \checkmark & \checkmark & \checkmark & \checkmark & {84.53 (+3.72\%)} & {66.06 (+39.19\%)} & {80.13 (+12.16\%)} \\
      \toprule
      \end{tabular}
      }}
      \vspace{-2em}
  \end{table*}

\begin{table}[t]
  \centering
      \caption{Ablation study on distillation loss in ConSept.}
      \vspace{-0.5em}
      \label{table:ablation_distill_loss}
      \small
      \setlength{\tabcolsep}{7.5pt} 
     \renewcommand{\arraystretch}{4.0}
   { \fontsize{8.3}{3}\selectfont{
      \begin{tabular}{cc|ccc}
      \bottomrule
      \multirow{2}{*}{\textbf{$\mathcal{L}_{\text{mse}}$}}&\multirow{2}{*}{\textbf{$\mathcal{L}_{\text{contrast}}$}}&\multicolumn{3}{c}{\textbf{15-1 (6 steps)}}
      \\ \cline{3-5}
      &&{{0-15}}&{{16-20}}&{all}\\ \hline
      - & - &81.81&64.35&77.65 \\
      \checkmark & - &84.35&63.77&79.45 \\
      \checkmark & \checkmark &84.53&66.06&80.13 \\
      \toprule
      \end{tabular}
      }}
      \vspace{-2em}
  \end{table}

\textbf{Effect of Distillation Loss}.
Subsequently, we investigate the impact of individual components in the distillation loss employed in ConSept. 
As depicted in Table~\ref{table:ablation_distill_loss}, ConSept achieves only 81.81\% in terms of base classes mIoU when $\mathcal{L}_{\text{mse}}$ and $\mathcal{L}_{\text{contrast}}$ are excluded. 
Upon introducing $\mathcal{L}_{\text{mse}}$, the performance improves to 84.35\% and 79.45\% in terms of mIoU on base classes and all classes, respectively. 
Finally, the inclusion of the auxiliary loss $\mathcal{L}_{\text{contrast}}$ results in an additional improvement of $\sim$2\% in novel classes mIoU. 
These findings suggest that introducing $\mathcal{L}_{\text{mse}}$ as the primary distillation loss contributes to the anti-catastrophic forgetting ability of base classes, 
while introducing auxiliary loss $\mathcal{L}_{\text{contrast}}$ enhances the generalization capability for novel classes.

\textbf{Distillation Strategy in ConSept}.
Given the observed improvement in performance resulting from the introduction of multi-scale features for segmentation, 
we are prompted to explore the applicability of  distillation loss across all scale features. 
To investigate the impact of candidate layers for distillation, an ablation study is conducted on ConSept with varying selections of distilled features. 
The corresponding quantitative results are presented in Table~\ref{table:ablation_distill_layers}. 
It is noteworthy that applying distillation loss solely on feature $\mathbf{F}^{3}_{t}$ from the deepest layers yields the optimal performance for both base and novel classes in ConSept. 
When extending the distillation loss to include features from shallow layers (\emph{i.e.}, $\mathbf{F}^{0-2}_{t}$), 
ConSept maintains similar performance in terms of mIoU on base classes, experiencing only a marginal performance decrease of less than 0.5\%. 
However, the performance on novel classes shows a notable degradation by $\sim$2\%. 
These results suggest that applying distillation loss exclusively to $\mathbf{F}^{3}_{t}$ is sufficient for ConSept. 
A plausible explanation is that networks should adapt to the texture and luminance information of novel classes from newly added data, 
necessitating the allowance for modification of features from regions associated with novel classes. 
Excessive distillation loss for $\mathbf{F}^{0-2}_{t}$ may impede the generalization ability for novel classes.

\begin{table}[t]
  \centering
      \caption{Ablation study on distilled feature selection in ConSept.}
      \label{table:ablation_distill_layers}
      \vspace{-0.5em}
      \small
      \setlength{\tabcolsep}{7.5pt} 
     \renewcommand{\arraystretch}{4.0}
   { \fontsize{8.3}{3}\selectfont{

      \begin{tabular}{c|ccc}
      \bottomrule
      \multirow{2}{*}{\textbf{Distilled Features}}&\multicolumn{3}{c}{\textbf{15-1 (6 steps)}}
      \\ \cline{2-4}
      &{{0-15}}&{{16-20}}&{all}
      \\ \hline
       $\mathbf{F}^{3}_{t}$ &84.53&66.06&80.13 \\
       $\mathbf{F}^{2-3}_{t}$ &84.53&63.99&79.64 \\
       $\mathbf{F}^{0-3}_{t}$ &83.99&63.84&79.19 \\
      \toprule
      \end{tabular}
      }}
      \vspace{-1em}
  \end{table}

  \begin{table}[t]
    \centering
        \caption{Ablation study on deterministic boundary in ConSept.}
        \label{table:ablation_linear_head}
        \small
        \setlength{\tabcolsep}{7.5pt} 
       \renewcommand{\arraystretch}{4.0}
     { \fontsize{8.3}{3}\selectfont{
  
        \begin{tabular}{c|ccc}
        \bottomrule
        \multirow{2}{*}{\textbf{Frozen Head}}&\multicolumn{3}{c}{\textbf{15-1 (6 steps)}}
        \\ \cline{2-4}
        &{{0-15}}&{{16-20}}&{all}
        \\ \hline
         - &81.79&66.34&78.11 \\
         \checkmark &84.53&66.06&80.13 \\
        \toprule
        \end{tabular}
        }}
        \vspace{-2em}
    \end{table}
\textbf{Deterministic Boundary or Fully Fine-tuning}.
Additionally, we seek to understand the impact of deterministic boundary on old classes. 
The frozen linear head for old classes intuitively ensures a well-defined decision boundary among classes, 
thereby preserving the segmentation capability of base classes and mitigating forgetting. 
To assess this effect, we compare the segmentation quality between ConSept with a frozen linear segmentation head for old classes and its counterpart, 
as outlined in Table~\ref{table:ablation_linear_head}. 
Freezing the linear head for old classes results in a 2.74\% improvement in base classes mIoU and a 2.01\% enhancement in overall mIoU for ConSept. 
Notably, the mIoU of novel classes remains unchanged. These findings suggest that freezing the linear head for old classes contributes to improving the anti-catastrophic forgetting capability for base classes, and consequently enhances the overall segmentation performance.

\begin{table}[t]
  \centering
      \caption{Ablation study on dual dice losses in ConSept.}
      \vspace{-1em}
      \label{table:ablation_dice_loss}
      \small
      \setlength{\tabcolsep}{7.5pt} 
     \renewcommand{\arraystretch}{4.0}
   { \fontsize{8.3}{3}\selectfont{
      \begin{tabular}{cc|ccc}
      \bottomrule
      \multirow{2}{*}{\textbf{$\mathcal{L}_{\text{c-dice}}$}}&\multirow{2}{*}{\textbf{$\mathcal{L}_{\text{on-dice}}$}}&\multicolumn{3}{c}{\textbf{15-1 (6 steps)}}
      \\ \cline{3-5}
      &&{{0-15}}&{{16-20}}&{all}
      \\ \hline
      - & - &80.31&64.54&76.55 \\
      \checkmark & - &81.86&64.71&77.76 \\
      \checkmark & \checkmark &84.53&66.06&80.13 \\
      \toprule
      \end{tabular}
      }}
      \vspace{-2em}
  \end{table}
  
\textbf{Effect of Dice Losses}.
Finally, we explore the impact of our proposed dual dice losses. 
The evaluation results are presented in Table~\ref{table:ablation_dice_loss}. 
Without incorporating the class-specific dice loss $\mathcal{L}_{\text{c-dice}}$ and the old-new dice loss $\mathcal{L}_{\text{on-dice}}$, ConSept achieves a base classes mIoU of 80.31\% and an overall mIoU of 76.55\%. 
Introducing $\mathcal{L}_{\text{c-dice}}$ results in a 1.55\% improvement in base classes mIoU with no impact on novel classes' performance. 
Further inclusion of $\mathcal{L}_{\text{on-dice}}$ leads to a notable performance boost of 2.37\% in overall mIoU and a 1.35\% enhancement in novel classes mIoU. 
These findings highlight that $\mathcal{L}_{\text{c-dice}}$ enhances segmentation quality for base classes during new tasks training, while $\mathcal{L}_{\text{on-dice}}$ further improves the segmentation ability for novel classes.


\section{Conclusion}\label{sec:conclusion}
In this paper, we investigated continual semantic segmentation with vision transformer. 
We empirically found that vanilla ViT inherently exhibits viable anti-catastrophic forgetting ability for base classes, and adapters could tackle this issue without extra negative effect. 
Therefore, we proposed, for the first time to our best knowledge, an adapter-based vision transformer for continual semantic segmentation tasks, namely \emph{ConSept}. 
Specifically, ConSept inserted lightweight attention-based adapters into pretrained ViT and adopted a dual-path architecture for segmentation. 
With less than 10\% additional parameters, ConSept obtained better segmentation ability for old classes, 
and achieved promising segmentation quality on novel classes. 
To further exploit the anti-catastrophic forgetting ability of ConSept, we proposed a distillation method with deterministic old-classes boundary for better anti-catastrophic forgetting ability, and dual dice losses to regularize segmentation maps for overall segmentation performance. 
Empirical quantitative and qualitative results illustrated that ConSept obtained new state-of-the-art performance on various continual semantic segmentation tasks. 
Meanwhile, ConSept demonstrated promising anti-catastrophic forgetting capability for both old classes and novel classes. 



{
\bibliographystyle{IEEEtran}
\bibliography{egbib}
}

\vfill

\end{document}